\documentclass[lettersize,journal]{IEEEtran}

\usepackage{booktabs}
\usepackage{multirow}
\usepackage[dvipsnames,table,xcdraw]{xcolor}
\usepackage{cite}
\usepackage{amsmath,amssymb,amsfonts}
\usepackage{bbm}
\usepackage{algorithm}
\usepackage{algorithmic}
\usepackage{color}
\usepackage{pbox}
\usepackage{graphicx}
\usepackage{url}
\usepackage{float}
\usepackage{graphbox}
\usepackage{enumitem}
\usepackage{multirow}
\usepackage{textcomp}
\usepackage{hyperref}
\usepackage{gensymb}
\usepackage{eurosym}
\usepackage{blindtext}
\usepackage{cellspace}
\usepackage[export]{adjustbox}
\usepackage{xcolor}
\usepackage{soul}
\soulregister\cite7
\usepackage[switch]{lineno}
\usepackage[hang,flushmargin]{footmisc}
\begin{document}

%\begin{frontmatter}


\title{Which cycling environment appears safer? Learning cycling safety perceptions from pairwise image comparisons}

\author{Miguel Costa$^{1,2,3}$, Manuel Marques$^{2}$, Carlos Lima Azevedo$^{3}$, Felix Wilhelm Siebert$^{3}$, and Filipe Moura$^{1}$% <-this % stops a space
%\thanks{}% <-this % stops a space
\thanks{\noindent This manuscript has been accepted for publication in IEEE Transactions on Intelligent Transportation Systems. © 2024 IEEE.  Personal use of this material is permitted.  Permission from IEEE must be obtained for all other uses, in any current or future media, including reprinting/republishing this material for advertising or promotional purposes, creating new collective works, for resale or redistribution to servers or lists, or reuse of any copyrighted component of this work in other works.}%
\thanks{$^{1}$
        Civil Engineering Research and Innovation for Sustainability, Instituto Superior T\'{e}cnico, Universidade de Lisboa, Av. Rovisco Pais, 1, Lisboa, Portugal,
        {\tt\footnotesize{fmoura@tecnico.ulisboa.pt}}}%
\thanks{$^{2}$
        Institute for Systems and Robotics, Instituto Superior T\'{e}cnico, Universidade de Lisboa, Av. Rovisco Pais, 1, Lisboa, Portugal, {\tt\footnotesize{manuel@isr.tecnico.ulisboa.pt}}}%
\thanks{$^{3}$
        Department of Technology, Management and Economics, Technical University of Denmark, Kgs. Lyngby, 2800, Denmark,
        {\tt\footnotesize{\{migcos, climaz, felix\}@dtu.dk}}}%
}

\maketitle

% % % % % % % % % % % % % % % % % % % % % % % % % % % % % % 
% % % % % %     Abstract & Keywords
% % % % % % % % % % % % % % % % % % % % % % % % % % % % % %
\begin{abstract}
Cycling is critical for cities to transition to more sustainable transport modes. Yet, safety concerns remain a critical deterrent for individuals to cycle. If individuals perceive an environment as unsafe for cycling, it is likely that they will prefer other means of transportation. Yet, capturing and understanding how individuals perceive cycling risk is complex and often slow, with researchers defaulting to traditional surveys and in-loco interviews. In this study, we tackle this problem. We base our approach on using pairwise comparisons of real-world images, repeatedly presenting respondents with pairs of road environments and asking them to select the one they perceive as safer for cycling, if any. Using the collected data, we train a siamese-convolutional neural network using a multi-loss framework that learns from individuals' responses, learns preferences directly from images, and includes ties (often discarded in the literature). Effectively, this model learns to predict human-style perceptions, evaluating which cycling environments are perceived as safer. Our model achieves good results, showcasing this approach has a real-life impact, such as improving interventions' effectiveness. Furthermore, it facilitates the continuous assessment of changing cycling environments, permitting short-term evaluations of measures to enhance perceived cycling safety. Finally, our method can be efficiently deployed in different locations with a growing number of openly available street-view images.

\end{abstract}


%\end{frontmatter}

%\linenumbers

% % % % % % % % % % % % % % % % % % % % % % % % % % % % % % 
% % % % % %     Main Text
% % % % % % % % % % % % % % % % % % % % % % % % % % % % % % 
\section{Introduction}
\label{sec:introduction}
% ######################################################
% CYCLING
% ######################################################
Cycling can promote short and long-term health benefits \cite{oja2011health, gotschi2016cycling} and reduce greenhouse gas emissions and air pollutants \cite{mason2015global, NEVES2019130}, compelling cities to promote cycling as a means of transportation. However, many deterrents exist for people not to cycle. These include social barriers~\cite{heinen2010commuting}, physical barriers~\cite{pucker2001cycling}, and psychological factors \cite{stinson2004frequency}.

% ######################################################
% SUBJECTIVE CYCLING SAFETY RESEARCH
% ######################################################
From these, safety concerns (fears of being involved in an accident) are often considered the main deterrent to cycling \cite{sanders2015perceived, felix2019maturing}. If individuals perceive cycling environments as unsafe, they will avoid cycling and most certainly prefer other modes of transportation. Previously, perception of safety has been found to be affected by different factors, including helmets and clothing \cite{lawson2013perception, aldred2015reframing}, road users' behavior \cite{chaurand2013cyclists}, road usage and compliance with rules \cite{lawson2013perception}, or infrastructure and cycling facilities \cite{moller2008cyclists, Wang2018, chataway2014safety, jensen2007road}. Research has been typically carried out using \textit{in situ} surveys or post-riding interviews and conventional choice modeling tools. However, these are often not scalable due to their inherent high costs, making them not easily redeployed over time or space.

% ######################################################
% NEW METHODS TO STUDY PERCEPTION OF CYCLING SAFETY
% ######################################################
% To circumvent this, new approaches have been proposed. These explore new ways of acquiring data, such as physiological data~\cite{zeile2016urban}, street-view imagery (SVI) about intersections~\cite{doorley2015analysis}, cyclists' perspective video clips~\cite{Parkin2007}, mental maps with geographical data \cite{manton2016using} or virtual reality~\cite{nazemi2019studying}. However, while these approaches seek more and new types of data on individuals' perceptions, most still require considerable manual labor to analyze and understand whether environments are perceived as safe or unsafe.

\begin{figure}[!tb]
\centering
%    \begin{minipage}[b]{.48\textwidth}
%        \centering
%        \begin{tabular}{c}
%            \includegraphics[width=.98\textwidth]{figs/survey_1.png}  \\
%        \end{tabular}
%    \end{minipage}
    \begin{minipage}[b]{.48\textwidth}
        \centering
        \begin{tabular}{c}
            \includegraphics[width=.98\textwidth]{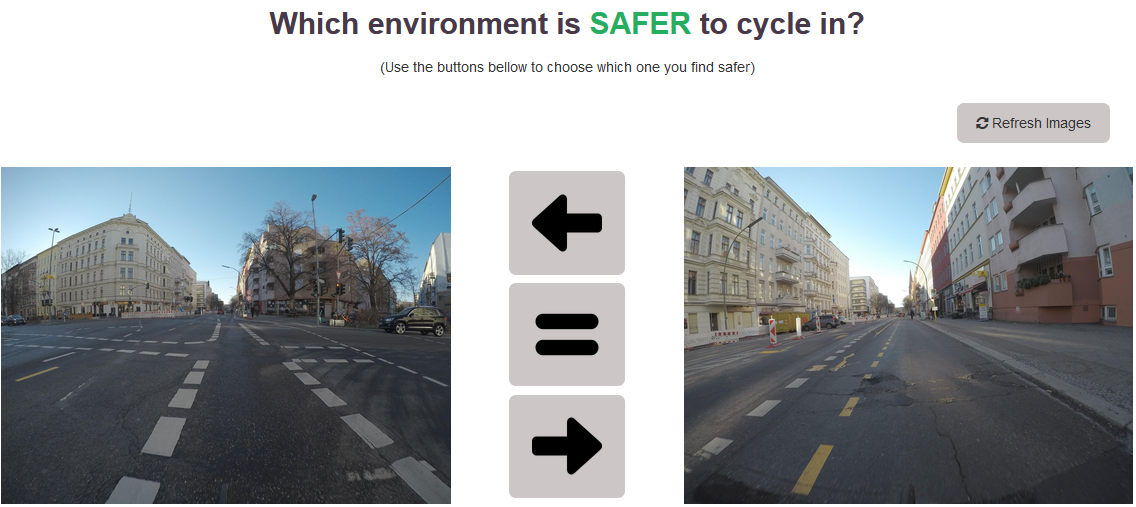}  \\
        \end{tabular}
    \end{minipage}
    \caption{Example of the pairwise image comparison survey. Users are shown two images and asked to select the one they consider safer for cycling, if any.}
    \label{fig:met_survey_example}
    \vspace{-12pt}
\end{figure}

% ######################################################
% Objectives & Contributions
% ######################################################
This work presents a novel methodological approach to more efficiently assess the perception of cycling safety, learning perceived cycling safety directly from real-world images and users' choices. Effectively, we want to answer the following question: ``\emph{Given two cycling environments, which is perceived as safer for cycling?}'', as shown in Figure~\ref{fig:met_survey_example} which presents an outline of the pairwise image comparison survey we conducted. Ultimately, we use the collected data to develop a model which can be used to score and map cycling environments' perceived safety, which we showcase for Berlin, Germany. Our work's main contributions are threefold:
\begin{itemize}[noitemsep,topsep=0pt]
    \item We create and deploy a new pairwise image comparison survey where we repeatedly present respondents with two road environment images and ask them to select the one they perceive as safer for cycling, if any.
    \item We propose PCS-Net, a neural network that predicts which environment is perceived as safer for cycling from two images trained on the collected comparisons.
    \item PCS-Net is able to predict and learn from ties (when there is no perceived difference between the two images), which are often discarded in past research.
\end{itemize}

% This work presents a novel methodological approach to efficiently assess the perception of cycling safety. Effectively, we want to answer the following question: ``\emph{Given two cycling environments, which is perceived as safer for cycling?}'' For this, we propose PCS-Net, a neural network that predicts which environment is perceived as safer for cycling when comparing two images. Moreover, PCS-Net is able to predict and learn from ties (when there is no perceived difference between the two images), which are often discarded in past research. To develop PCS-Net, we create and deploy a new pairwise image comparison survey where we repeatedly present respondents with two road environments and ask them to select the one they perceive as safer for cycling. In effect, PCS-Net learns the perception of cycling safety directly from two real-world images and users' choices. Ultimately, this model can be used to score perceived safety for different cycling environments, effectively allowing one to map perceived safety geographically, which we showcase for Berlin, Germany. 

% ######################################################
% Outline of article
% ######################################################
Next, we overview related work in Section~\ref{sec:related_work}. Section~\ref{sec:methods} explains our approach, including data and models. Experimental results are detailed in Section~\ref{sec:results}, and we apply PCS-Net at a city-wide scale in Section~\ref{sec:application}. Section~\ref{sec:discussion} discusses our results and its limitations. Section~\ref{sec:conclusions} finalizes the paper and draws possible paths for future research.

\section{Related Work}
\label{sec:related_work}
% ######################################################
% SUBJECTIVE CYCLING SAFETY
% ######################################################
\subsection{Subjective Cycling Safety}

% ######################################################
% OVERALL
% ######################################################
Perceived or subjective cycling safety relates to how humans subjectively experience cycling accident risk. As the most critical deterrent to urban cycling \cite{sanders2015perceived, felix2019maturing}, understanding what impacts it is vital to adequately provide cyclists with environments they feel safe. Aspects that contribute to this sense of safety include: traffic \cite{sanders2015perceived}, interactions with drivers and road rules compliance \cite{lawson2013perception, GRAYSTONE2022103237}, sharing the road space \cite{pyrialakou2020perceptions}, and road and cycling infrastructure \cite{chataway2014safety, NG201713}.

% ######################################################
% METHODS TO STUDY SUBJECTIVE CYCLING SAFETY
% ######################################################
Past research has typically focused on qualitative surveys, \textit{in situ} or post-riding interviews to identify elements that negatively arouse individuals \cite{sanders2015perceived, aldred2015reframing}. However, these are often time-consuming and costly, requiring critical preparation and control, making these approaches not scalable nor transferable between cities. Additionally, due to their inherent focus on specific routes or infrastructure typologies, recruitment of interviewees may not be easy, leading to non-diverse and non-representative groups of individuals being interviewed \cite{bosen2023cycling}. 

More recent approaches, mainly driven by technological advances, have led researchers to conduct other experiments. These focus on quantitatively assessing human responses, including using wearable sensors~\cite{zeile2016urban}, cycling videos~\cite{Parkin2007}, drawings of mental maps~\cite{manton2016using}, virtual reality~\cite{von2018risk}, semi-realistic street-view-style images~\cite{von2022safe}, or eye tracking devices~\cite{schmidt2018risk}. Yet, these usually require precise calibration and monitoring or individual training to use such technology, hindering their transferability and repeatability.

% ######################################################
% SUBJECTIVE CYCLING SAFETY INDICES
% ######################################################
All in all, understanding the impact of different elements on individuals' perceptions can lead to objectively characterize urban environments, leading to the creation of indicators or indices that can help urban planners compare and analyze different environments. To this end, the Bicycle Stress Level \cite{sorton1994bicycle} and the Level of Traffic Stress \cite{mekuria2012low, furth2017level} remain the most well-known indices that attempt to measure bikeability and perceived risk. Yet, to compute such metrics, manual labor is usually employed, requiring individuals to annotate environment elements manually. \cite{ito2021assessing} approach fills this gap, deriving a bikeability index from computer vision-extracted features from street-view images (SVI). This automatic and scalable methodology to score environments effortlessly can eventually replace more traditional techniques. Nevertheless, it covers five bikeability aspects, of which perceptions are one, which can be inadvertently mischaracterized if appropriate supervision is not employed. In this work, we are interested in a similar approach that covers the perception of cycling risk only, is based on individuals' perceptions of what categorizes one environment as safer than another, and is equally scalable. Such an approach, in principle, can simulate individuals' perceptions and thus reduce the time and cost of assessing perceived safety compared to traditional methodologies.

\vspace{-6pt}
% ######################################################
% COMPUTER VISION, STREET-VIEW IMAGERY & URBAN STUDIES 
% ######################################################
\subsection{Computer Vision \& SVI applied in Urban Studies}

Recently, a growing number of works have been exploring image processing and computer vision techniques to study urban environments and human perceptions. For example, these have explored openness and enclosure~\cite{LI201881}, greenery~\cite{toikka2020green}, house prices~\cite{nouriani2022vision}, and different human perceptions~\cite{naik2014streetscore, dubey2016deep, verma2020predicting, RAMIREZ2021104002, guan2021urban}.

Research has typically used traditional or deep learning-extracted features from SVI in classification or regression problems to evaluate the impact of such environment-related features. With the growing access to openly available street-view images, researchers can perform continuous and scalable assessments of urban environments more easily than in the past. In transportation, using SVI has been proven useful in, among others, studying accessibility~\cite{najafizadeh2018feasibility, saha2019project}, walkability~\cite{RAMIREZ2021104002}, bikeability~\cite{ito2021assessing}, and objective road safety~\cite{song2018farsa}. 

With this in mind, our work is greatly inspired by that of \cite{dubey2016deep}, which uses computer vision to predict how individuals sense urban environments across different perspectives. The authors use a large database of image comparisons to train a model capable of "simulating" human choice. Such work has laid the foundations for many possible city-wide analyses, such as understanding how the built environment might affect behavior, travel choices, and even home location. Further, as the authors demonstrate and hypothesize, such an approach can be used to enlarge existing datasets and expand predictions across geographies. This endeavor can help to better allocate city resources and make data-driven decisions. Yet, their model disregards a vast number of observations as it cannot handle ties (equally perceived images). We consider that there is a lot of information to be gained here. Thus, we build up from \cite{dubey2016deep}'s approach, proposing a methodology that draws from pairwise image comparisons, includes ties, and can be applied to analyze cycling perception of safety.

\vspace{-6pt}
% ######################################################
% PAIRWISE COMPARISONS
% ######################################################
\subsection{Pairwise Comparisons}
Pairwise comparison models aim to predict the outcome of a two-item comparison, i.e., when comparing $A$ and $B$, would a user prefer $A$, $B$, or would they be perceived equally (tie)? Models typically follow the seminal works of \cite{thurstone1927law} and \cite{bradley1952rank}, which assume a latent score $s$ and the outcome probability of comparing items $i$ and $j$ a function of their scores, e.g., a user would likely choose item $i$ if $s_i>s_j$. The underlying goal is then to estimate the latent scores $s_i$ from data.

In principle, this is similar to the approach often used in contrast learning or rank sorting. For one, contrast and comparison in the perception of human beings has led to contrastive learning method \cite{ding2023end} where a model learns similarities and differences by looking at positive and negative examples versus an original example. Here, the goal is to maximize similarities between identical items while minimizing similarities between dissimilar items. On the other hand, in rank sorting, the goal is to order items based on a learnt latent score (rank). Yet, both approaches, together with paired models, seek to objectively uncover an item's latent score that can be used later to understand how similar or dissimilar items are.

In paired models, several methodologies have been proposed to extend the mentioned seminal works, including iterative algorithms \cite{elo1978rating}, Bayesian models \cite{herbrich2006trueskill}, spectral ranking algorithms \cite{chau2023spectral}, convex problem formulation \cite{xu2016pairwise}, Gaussian processes \cite{maystre2019pairwise}, or other deep learning approaches \cite{engilberge2019sodeep}. Yet, most of these do not consider an item's characteristics. In this work, we focus on deriving items' scores directly from their features, and we hypothesize that including such information can help achieve better prediction accuracy.

\section{Learning from pairwise image comparisons}
\label{sec:methods}
Images are vital in conveying messages and representations of real-world. We frame our approach under this context, using real-world images to map users' safety perceptions of cycling environments. In this section, we will now detail our approach, including data collection and model training to predict which cycling environments are perceived as safer for cycling. 

\vspace{-6pt}
% ######################################################
% DATA: SURVEY
% ######################################################
\subsection{Data: Pairwise Image Comparisons Survey}
\label{sec:methods_data}

% First, we created a survey that allows us to capture data about individuals' safety perceptions semi-naturalistically. The result is a two-part survey that was approved by Instituto Superior Técnico's Ethics Committee and i) captured individuals' cycling profiles (i.e., cycling proficiency), and ii) perceived cycling accident safety in different environments using pairwise image comparisons. In this work, we focus exclusively on the second part. The survey was deployed online and took about 10-15 minutes to complete. Overall, 251 individuals partook in the survey, completing 7281 paired comparisons (3.3 average comparisons per image), of which 18\% were ties.

First, we created a survey to capture data about individuals' safety perceptions semi-naturalistically.
This two-part survey was approved by Instituto Superior T\'ecnico's Ethics Committee and i) captured individuals' cycling profiles (i.e., cycling proficiency), and ii) perceived cycling accident safety in different environments using pairwise image comparisons.
The first part consists of a modified version of \cite{dill2016revisiting} that classified individuals according to four cycling profiles: \textit{No Way, No How} (NWNH), \textit{Interested but Concerned} (IC), \textit{Enthused and Confident} (EC), and \textit{Strong and Fearless} (SF) \cite{geller2006four}. The socio-demographic characteristics of participants were also captured. 
The survey was deployed online and took about 10-15 minutes to complete. Overall, 251 individuals partook in the survey.
These were characterized as being 6\% NWNH, 53\% IC, 37\% EC, and 4\% SF; 60\% identified as male, 40\% as female, and $<$1\% as other/prefer not to answer (NA); $<$1\% were aged 18--20, 45\% were 21--30, 30\% were 31--40, 16\% were 41--50, 8\% were 51--60, and the rest were 61, older or NA.

% ######################################################
% PAIRWISE IMAGE COMPARISONS
% ######################################################
In this work, we focus exclusively on the second part of the survey. Data collected consists of 7281 pairwise image comparisons (3.3 average comparisons per image), of which 18\% were ties. Unlike traditional surveys, pairwise comparisons are straightforward, well suited for non-expert participants, and generally lead to lower measurement error than direct ratings~\cite{perez2017practical}. We repeatedly presented respondents with two real-world images of cycling environments and asked them to select the one they thought was safer for cycling. Equally sized images were shown side-by-side and users could choose between the left image ($y=-1$), right image ($y=1$), or, if they thought there was no apparent difference between the two, they could choose a ``tie'' ($y=0$) option. Figure~\ref{fig:met_survey_example} depicts the image comparison setup.

To choose what pairs of images were shown to participants, we used an iterative approach. We began by downloading 4,480 road environment pictures of Berlin, Germany, from Mapillary (\url{https://www.mapillary.com/}), capturing various infrastructure layouts, urban features, lighting conditions, and traffic conditions. Given the diversity of cycling environments in these images, we employed a fractional factorial design to select which pairs of images to present to users. 
Each image and its corresponding cycling environment were pre-processed through semantic segmentation and built environment data extraction. We applied OCRNET \mbox{\cite{yuan1909segmentation}} to segment each image and identify object areas, such as vegetation, people, and cars. In total 19 classes of objects are extracted. Additionally, we extracted built environment data from OpenStreetMaps (\url{https://www.openstreetmap.org/}, OSM), including road hierarchies, land use, presence of cycling lanes, and urban furniture. In total, 192 urban elements are captured from OSM.

Using both the semantic segmentation data and street-level data, we created potential image pairs by matching specific information. To accomplish this, we begin by randomly selecting one image and randomly choosing eight possible environment characteristics (e.g., cars, vegetation, sidewalks, pedestrians, primary roads, cycling lanes, shops, traffic lights) from the set of 211 possible variables. 
Next, we look for other images that contain similar values of such characteristics, while allowing other variables to vary. 
After selecting the two images, we presented the pair to the participant. This fractional factorial design maximizes the information obtained from image pairs while reducing the number of experimental runs required, which would be unmanageable in a full factorial design. Finally, to ensure balanced exposure, we prioritized showing images that had been displayed fewer times in the past so that all images were shown an equal number of times.

\begin{figure*}[htb]
    \centering
    \begin{minipage}[b]{.65\textwidth}
        \centering
        \includegraphics[trim=0 95 0 0,clip, width=.99\textwidth]{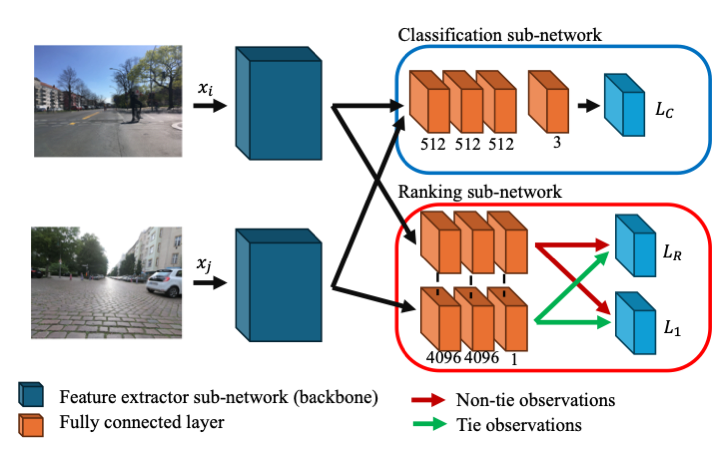}  \\
    \end{minipage}
    \begin{minipage}[b]{.25\textwidth}
        \centering
        \includegraphics[trim=0 150 500 150,clip, width=.99\textwidth]{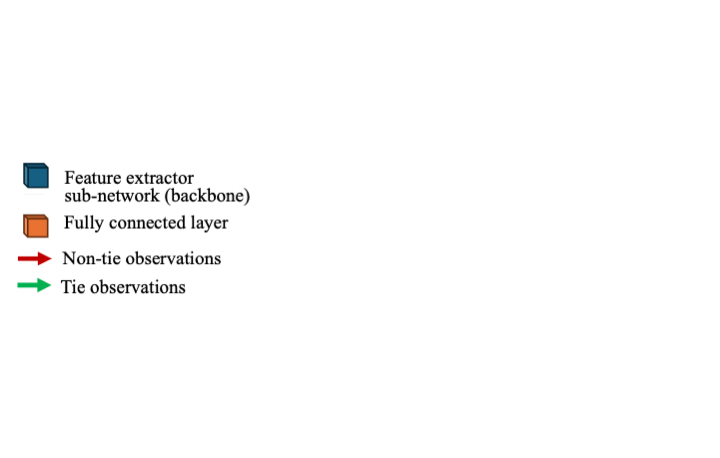}  \\
    \end{minipage}\\
    \vspace{-8pt}
  \caption{Architectures of PCS-Net$_C$ (above) and PCS-Net$_R$ (below) and corresponding losses ($L_C$ for PCS-Net$_C$ and $L_R$ and $L_1$ for PCS-Net$_R$), which, when combined, make up PCS-Net. Layer sizes are shown below each layer.}
  \label{fig:met_network_architecture}
  \vspace{-12pt}
\end{figure*}
%\begin{figure*}[!htb]
%  \centering
%  \includegraphics[width=.65\textwidth]{figs/network_1.png}  \\
%  \caption{Architectures of the PCS-Net$_C$ (above) and PCS-Net$_R$ (below) networks, which, when combined, make up PCS-Net.}
%  \label{fig:met_network_architecture}
%  \vspace{-12pt}
%\end{figure*}

% ######################################################
% Multi-loss siamese-convolutional model
% ######################################################
\subsection{PCS-Net}
Our model seeks to uncover the underlying patterns and principles that guide individuals' subjective safety judgments when comparing and selecting between different images. Specifically, we seek to learn the transformation function $f(x_i)$ to predict an individual's choice $y$ when presented with images $x_i$ and $x_j$. The models outlined here follow the rationale of the networks by \cite{dubey2016deep}, namely their classification and ranking networks. However, we modify these networks to better suit our problem and expand them to allow ties in comparisons as we find these can often bring information gain. 

% ######################################################
% CLASSIFICATION
% ######################################################
\noindent\textbf{Classification Problem:} Given an individual's choice $y\in\{-1, 0, +1\}$, we want our model to predict said choice, i.e., the user chose the left image, a tie, or the right image. As such, we begin by designing the Classification-Perception of Cycling Safety Network (PCS-Net$_C$). Figure~\ref{fig:met_network_architecture} shows the architecture of this network, which is divided into two sections: i) a feature extractor sub-network (backbone) and ii) a fusion network. The former is a Siamese-style network \cite{chopra2005learning}, where each identical branch with tied weights learns to map an input image to a latent representation of itself, acting as a feature extractor. Next, both branches are merged in the fusion sub-network \cite{dubey2016deep}, which in turn learns a combination of filters, ending with a softmax loss ($L_C$) used to train the network.

% ######################################################
% RANKING
% ######################################################
\noindent\textbf{Ranking Problem:} The main goal is to find a ranking score for each image, such that it represents an ordinal ranking among all images in the dataset. To make a prediction, ranking scores are compared between the two input images, and a choice is made following the highest-ranked image, i.e., the one perceived as safer. To achieve this, a slightly different network is designed: Ranking-Perception of Cycling Safety Network (PCS-Net$_R$), as shown in Figure~\ref{fig:met_network_architecture}. This network is subdivided into two components: i) a feature extractor sub-network (equal to the classification formulation) and ii) a ranking sub-network \cite{dubey2016deep}. In essence, this sub-network uses fully connected layers to reduce the features extracted from the backbone to a ranking score for an input image. As such, we learn function $f(x_i)$, such that when a choice $y$ is made between $x_i$ and $x_j$ we want to satisfy
\begin{equation}
-y \cdot (f(x_i) - f(x_j)) > 0,
\end{equation}
with $y\in\{-1, +1\}$, with $y=-1$ for when image $x_i$ wins the comparison, and $+1$ if $x_j$ wins. In essence, $f(x)$ ranks images based on their features, with higher-ranked images denoting images perceived safer for cycling. This problem can be expanded to denote a ranking loss:
\begin{equation}
L_{\hat{R}}(x_i, x_j) = \max(0, -y \cdot (f(x_i) - f(x_j)) + \gamma),
\end{equation}
which also allows the introduction of a margin term $\gamma$. Using this loss function, we are essentially penalizing comparisons in which image ranking orders are opposite to the individual's choice while favoring comparisons where a choice was made for an image ranked higher (perceived safer). However, $L_{\hat{R}}$ does not account for possible ties in a decision maker's choice ($y=0$). When a tie occurs, perceived cycling safety scores should be similar, as the respondent could not distinguish which one was deemed safer between the two images. This entails that both image rankings should be closer, which can be translated using a different loss for tie comparisons:
\begin{equation}
L_1(x_i, x_j) = \max(0, ||f(x_i) - f(x_j)||_1 - \gamma),
\end{equation}
with $|\cdot|_1$ being the L1 norm. In essence, $L_1$ pushes images from tie comparisons together. Finally, to train our model, we combine both losses and minimize:
\begin{equation}
L_R = \mathbbm{1}_{y\in\{-1,+1\}} L_{\hat{R}} + \lambda_1 \cdot \mathbbm{1}_{y\in\{0\}} L_1,
\label{eq:ranking_loss}
\end{equation}
with $\mathbbm{1}_{y\in\{-1,+1\}}$ being an indicator function for non-tie comparisons, $\mathbbm{1}_{y\in\{0\}}$ an indicator function for tie comparisons, and $\lambda_1$ a weight term to modulate the importance of ties.

% ######################################################
% CLASSIFICATION + RANKING
% ######################################################
\noindent\textbf{Classification \& Ranking Problem:} Finally, we design an approach that combines both the classification and ranking tasks together. This approach learns in an end-to-end framework how to predict which of the two input images is considered safer for cycling. This Perception of Cycling Safety Network (PCS-Net) joins all the above architectures: i) feature extractor, ii) fusion, and iii) ranking sub-networks. Training is performed by minimizing the multi-loss function:
\begin{equation}
L =  L_C + \lambda_{\hat{R}} \cdot \mathbbm{1}_{y\in\{-1,+1\}} L_{\hat{R}} + \lambda_1 \cdot \mathbbm{1}_{y\in\{0\}} L_1,
\label{eq:global_loss}
\end{equation}
with $\lambda_{\hat{R}}$ and $\lambda_1$ being hyper-parameters to be chosen to model loss component importance and maximize model accuracy. We experiment with different hyper-parameters in Section~\ref{sec:results}.

\begin{figure*}[!htb]
    \centering
    \includegraphics[width=.9\textwidth]{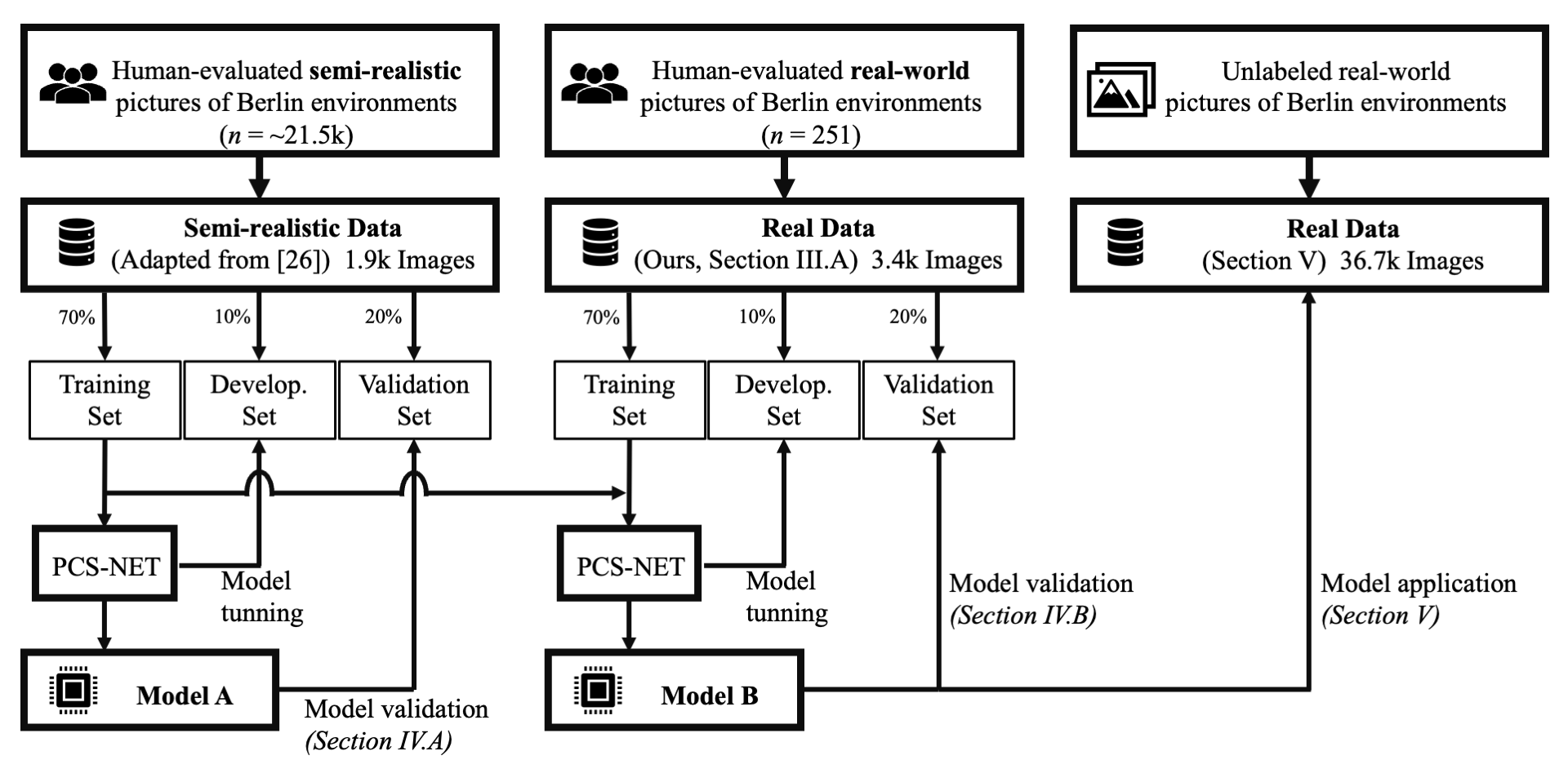}  \\
    \vspace{-8pt}
  \caption{Framework used for model validation using human labelled data (semi-realistic and our real-world data). Data is split in training sets to train Models A and B, which are evaluated on the development sets until convergence is met. Models A and B are then validated on unseen data (validation sets). The best performing model is then used for extrapolating perceived scores for unlabelled images at a city-wide scale in Section~\ref{sec:application}.}
  \label{fig:met_framework}
  \vspace{-6pt}
\end{figure*}
%\begin{figure*}[!htb]
%  \centering
%  \includegraphics[width=.65\textwidth]{figs/network_1.png}  \\
%  \caption{Architectures of the PCS-Net$_C$ (above) and PCS-Net$_R$ (below) networks, which, when combined, make up PCS-Net.}
%  \label{fig:met_network_architecture}
%  \vspace{-12pt}
%\end{figure*}

\begin{figure*}[!htb]
\begin{minipage}[b]{.90\textwidth}
    \centering
    \begin{tabular}{c}
        \includegraphics[width=.98\textwidth]{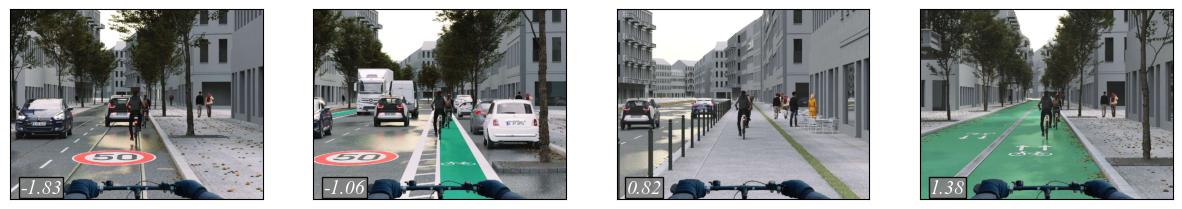}  \\
    \end{tabular}
\end{minipage}
\begin{minipage}[b]{.90\textwidth}
    \centering
    \begin{tabular}{c}
        \includegraphics[width=.98\textwidth]{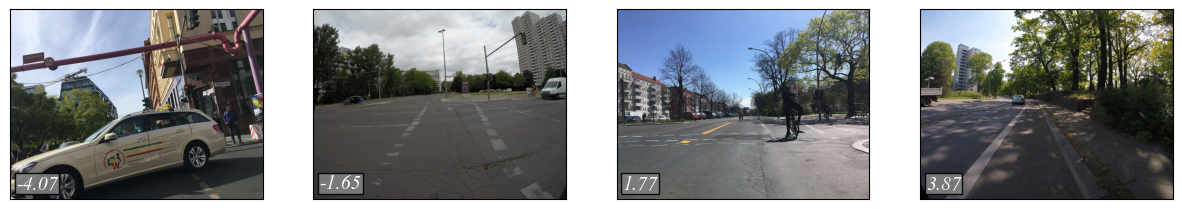}  \\
    \end{tabular}
\end{minipage}
\centering
  \vspace{-6pt}
  \caption{Examples of images ranked by their perceived safety scores. Ranked semi-realistic from Model (top) A and real images from Model B (bottom) with increasing (left to right) perceived safety scores are shown.}
  \label{fig:res_ranked_images}
  \vspace{-6pt}
\end{figure*}

\section{Results}
\label{sec:results}
% ######################################################
% INTRO
% ######################################################

We now present the results from our work. Figure \ref{fig:met_framework} shows our model training and validation framework. To showcase PCS-Net's ability to capture perceived cycling safety from images, we use two datasets originating from human-evaluated cycling environment images (in addition to our own dataset described in Section \ref{sec:methods_data}, we use data adapted from \cite{von2022safe}) and estimate two models. In either dataset, we randomly split pairwise image comparisons in 70-10-20\% for training, development, and validation (hold-out, test set). 

As typically done, we train both models using the training set and iteratively test them on the development set to uncover the best hyperparameters that produce the best model. The best-performing model is then tested and validated on the validation set containing observations the model has not seen before. All results below pertain to the models' results on the validation set. Unless otherwise stated, we compare our approach to other paired models that do not include ties using accuracy. We compute a non-margin accuracy metric for non-tie observations as the ratio of the correctly predicted left and right comparisons to all comparisons.

All experiments used Python, PyTorch2 \cite{pytorch}, and an NVIDIA GeForce 3080Ti GPU. Weights for the ranking and fusion sub-network were initialized from a uniform distribution relative to each layer's size. Learning rate was set to $0.001$, decaying every 10k steps, ADAM \cite{kingma2014adam} as our optimization procedure, and a batch size of 128 and up to 20 maximum epochs, or until validation accuracy stopped improving. We make all data and code publicly available online: \texttt{\url{https://github.com/mncosta/cycling_safety_subjective_learning_pairwise}}.

% ######################################################
% TESTING ON SEMI-REALISTIC DATA
% ######################################################
\vspace{-12pt}
\subsection{Model A}
% \subsection{Testing on semi-realistic data}

To validate our proposed model, we begin by using a dataset of semi-realistic images of 1900 computer-generated infrastructure typologies in Berlin \cite{von2022safe}. The original data contains answers to an online survey where users were asked to assess their perception of safety using a 4-point scale. We adapted this dataset to our use case, transforming each judged image to pairwise comparisons following \cite{dittrich2007paired}'s approach. For each user, if image A was judged higher than B, we generated a comparison between A and B, where A was chosen. If both images were scored equally, then we generated a tie. We repeated this process for all users, resulting in $\sim$2 million pairwise comparisons. We use these comparisons and images to train PCS-Net. We name the fitted model as Model A.

Figure~\ref{fig:res_ranked_images} showcases examples of images ranked by Model A, together with increasing (left to right) perceived safety scores. The left-most image depicts the lowest perceived environment, where a cyclist shares a road with tram rails and other traffic. On the opposite side, we can see a cyclist on a dedicated cycle lane as being perceived as very safe.

\begin{figure}[h]
    \centering
    \begin{minipage}{0.45\textwidth}
        \includegraphics[width=.98\textwidth]{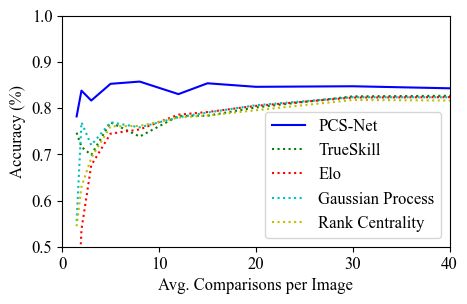}  \\
    \end{minipage}
    \vspace{-21pt}
    \caption{Model accuracy with varying number of average comparisons for different paired models.}
    \label{fig:res_synthetic_pc}
    \vspace{-12pt}
\end{figure}

We employ such a dataset given its vast amount of observations, allowing us to compare our model to other pairwise comparison methods: TrueSkill~\cite{herbrich2006trueskill}, Elo~\cite{elo1978rating}, Gaussian Process~\cite{maystre2019pairwise}, and Rank Centrality~\cite{negahban2012iterative}. Table~\ref{tab:models_results_benchmark_synthetic} shows the accuracy of our model versus other models, achieving comparable performance for the ranking models and a subpar performance for the classification task. Now, suppose we reduce the number of available average pairwise comparisons per image (effectively reducing the number of available comparisons). When constrained by the size of the training data, our model vastly outperforms the remaining pairwise models, as shown in Figure~\ref{fig:res_synthetic_pc}, reaching about 15\% improved accuracy, which can be very beneficial when data acquired from surveys is low.

\begin{table}[!htb]
\centering
\caption{Comparison of Model A versus other paired comparison models.}
\label{tab:models_results_benchmark_synthetic}
\vspace{-6pt}
\begin{tabular}{lrr}
Model    & Accuracy $\uparrow$\\
\hline \hline
Elo                   \cite{elo1978rating}         &  0.833  \\
TrueSkill             \cite{herbrich2006trueskill} &  0.839 \\
Gaussian Process      \cite{maystre2019pairwise}   &  0.843  \\
Rank Centrality       \cite{negahban2012iterative} &  0.839  \\
\hline
PCS-Net$_C$ [VGG] (Ours)                           &  0.599  \\
PCS-Net$_R$ [VGG] (Ours)                           &  0.839  \\
PCS-Net [VGG] (Ours)                               &  0.839  \\
\hline
\end{tabular}
\vspace{-12pt}
\end{table}

\begin{table}[!htb]
\centering
\caption{Comparison of Model B versus other paired comparison models.}
\label{tab:models_results_benchmark}
\vspace{-6pt}
\begin{tabular}{lrr}
Model    & Accuracy $\uparrow$\\
\hline \hline
Elo                   \cite{elo1978rating}         &  0.591  \\
TrueSkill             \cite{herbrich2006trueskill} &  0.624  \\
Gaussian Process      \cite{maystre2019pairwise}   &  0.632  \\
Rank Centrality       \cite{negahban2012iterative} &  0.611  \\ \hline
PCS-Net$_C$ [VGG] (Ours)                           &  0.849  \\
PCS-Net$_R$ [VGG] (Ours)                           &  0.855  \\
PCS-Net [VGG] (Ours)                               &  \textbf{0.867}  \\
\hline
\end{tabular}
\vspace{-12pt}
\end{table}

\begin{table}[!htb]
\centering
\caption{Model B accuracy using different backbones, with and without data augmentation using semi-realistic data.}
\label{tab:backbone_metrics}
\begin{tabular}{llr}
\begin{tabular}[c]{@{}l@{}}Augmented with \\ Semi-realistic Data\end{tabular} & Backbone   & Accuracy $\uparrow$\\ \hline
\hline
\multirow{4}{*}{No}  &    AlexNet~\cite{NIPS2012_c399862d} &    0.832   \\ \cline{2-3}
                     &        VGG~\cite{simonyan2014very}  &    0.867   \\ \cline{2-3} 
                     &     ResNet~\cite{he2015deep}        &    0.846   \\ \hline
\multirow{4}{*}{Yes} &    AlexNet~\cite{NIPS2012_c399862d} &    0.855   \\ \cline{2-3} 
                     &        VGG~\cite{simonyan2014very}  &    \textbf{0.874}   \\ \cline{2-3} 
                     &     ResNet~\cite{he2015deep}        &    0.846   \\ \hline
\end{tabular}
\vspace{-12pt}
\end{table}

% ######################################################
% TESTING ON REAL DATA
% ######################################################
\subsection{Model B}
% \\subsection{Testing on real data}

Next, we move to our work's core results, which apply PCS-Net to real-world images and pairwise comparisons, as detailed in Section~\ref{sec:methods_data}. Data contains responses from 251 users on images from Berlin on 7281 comparisons (3.3 average comparisons per image), of which 18\% consist of ties.

\begin{figure*}
\begin{minipage}{0.49\textwidth}
    \input{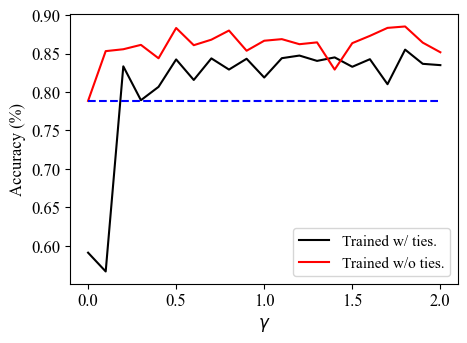}
\end{minipage}\hfill
\begin{minipage}{0.49\textwidth}
    \input{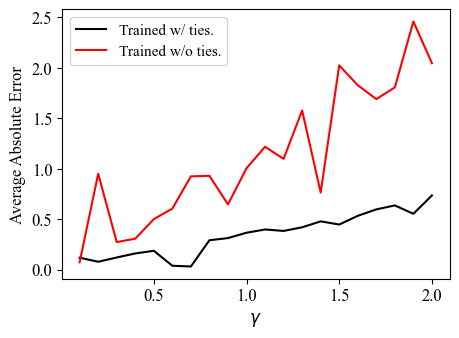}
\end{minipage}
\vspace{-14pt}
\end{figure*}

\begin{figure*}
\begin{minipage}{0.49\textwidth}
    \input{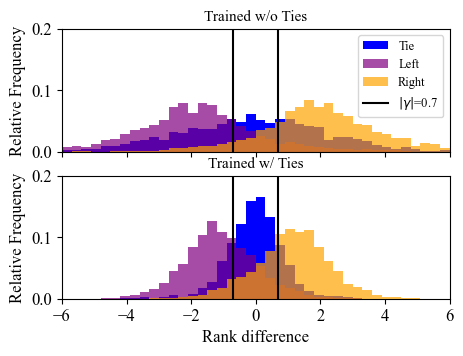}
\end{minipage}\hfill
\begin{minipage}{0.49\textwidth}
    \input{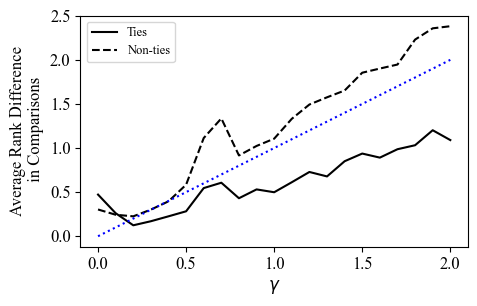}
\end{minipage}\hfill
\vspace{-14pt}
\end{figure*}

%\input{figs/all_accuracy_2}

% ##################################
% PAIRED MODELS
We begin by comparing our approach versus other pairwise models to compare their performance in predicting which cycling environment appears safer. Again, we test our method against TrueSkill~\cite{herbrich2006trueskill}, Elo~\cite{elo1978rating}, Gaussian Process~\cite{maystre2019pairwise}, and Rank Centrality~\cite{negahban2012iterative}, shown in Table~\ref{tab:models_results_benchmark}. As expected, given the low number of average pairwise comparisons per image available in our data, PCS-Net outperforms competing approaches with a $\sim$17\% improvement. This means that PCS-Net can effectively learn rankings directly from images, even when the number of available comparisons is limited.

%\input{tables/paired_models}

% ##################################
% BACKBONES
Next, we experiment with different network backbones (feature extraction sub-network) and different model hyperparameters. Backbone networks (feature extractor sub-networks) were initialized using the pre-trained Imagenet weights available publicly: AlexNet~\cite{NIPS2012_c399862d}, VGG~\cite{simonyan2014very}, and ResNet~\cite{he2015deep}. Table~\ref{tab:backbone_metrics} shows the model's accuracy using different backbones. VGG achieves the highest accuracy, and it is closely followed by ResNet and AlexNet. Additionally, we test whether augmenting our real image dataset by adding semi-realistic data improves model performance. For this, we train our model using a combination of real-world images and a similar-sized set of random semi-realistic pictures. Looking at the lower section of Table~\ref{tab:backbone_metrics}, we see that results improve slightly, with the highest accuracy being achieved again with VGG. Thus, augmenting our dataset with semi-realistic image comparisons leads to better prediction accuracy.

% ##################################
% MARGINS
Next, we test the impact of including ties and use of the margin $\gamma$. Figure~\ref{fig:res_accuracy_nomargin} shows the model's accuracy over $\gamma$ when PCS-Net is trained with and without ties. Additionally, we plot the default baseline when PCS-Net does not account for ties or margin effects (in blue) for comparison. When we set $\gamma>0.4$, the model trained with ties achieves an average 3\% lower accuracy than the non-tie trained model's accuracy. 

So far, we have looked at model performance and compared them to approaches that either do not allow ties or do not include such additions easily. Yet, including ties allows participants not to choose a preference when, in fact, there is none. Moreover, knowing when there is no distinguishable difference between two images can be highly valuable in practice, informing planners and decision-makers which environments are perceived equally. To account for this aspect, we now compute a different accuracy metric to include \textit{ties} as an output class and use $\gamma$ to distinguish between ties and non-ties. This new metric is computed as the ratio of the correctly predicted left (i.e., $f(x_i) > f(x_j) + \gamma$), right ($f(x_j) > f(x_i) + \gamma$), and tie ($|f(x_i)-f(x_j)| < \gamma$) comparisons to all comparisons. It concedes an interval defined by $\gamma$ when there is insufficient rank difference between images. When this happens, we consider there is no noticeable difference between the two images, and the comparison is expected to result in a tie. Using this new, more general metric, one can rank images and simultaneously test whether a comparison is expected to result in a tie. We again test our models trained with and without ties and compute the new (3-class) accuracy and error rates for $\gamma$, achieving comparable results.
However, analyzing the magnitude error of incorrectly classified observations (i.e., loss of misclassified observation) in Figure~\ref{fig:res_error_misclassified_margin}, the model trained with ties achieves an average absolute error much lower than its non-ties trained counterpart. Moreover, the absolute error on wrongly classified images was relatively low ($<0.07$) at its minimum ($\gamma=0.7$), suggesting that even for misclassified comparisons, including ties leads to a lower error.

Using this $\gamma$, one can also analyze the distribution of image rank differences for comparisons between a model trained with and without ties. Looking at Figure~\ref{fig:res_rank_difference_distribution}, the model incorporating ties (below) pushes ties below margin $|\gamma|$ and non-ties above $|\gamma|$. In the model trained without ties, ties are much less distinct and are dispersed among the left and right options. Additionally, Figure~\ref{fig:res_rank_difference} showcases the average rank differences in comparisons for different margins. Notably, the average difference for ties lies below $\gamma$, whereas above $\gamma$ for non-ties. This approach corroborates the idea of allowing participants to opt for ties to better reproduce their perceptions and improve their engagement in comparing images more carefully, as they are not forced to choose an option they are slightly unsure of.

\section{City-wide application}
\label{sec:application}
\begin{figure*}[!htb]
\hspace*{\fill}
\begin{minipage}[b]{.83\textwidth}
    \centering
    \includegraphics[width=.95\textwidth]{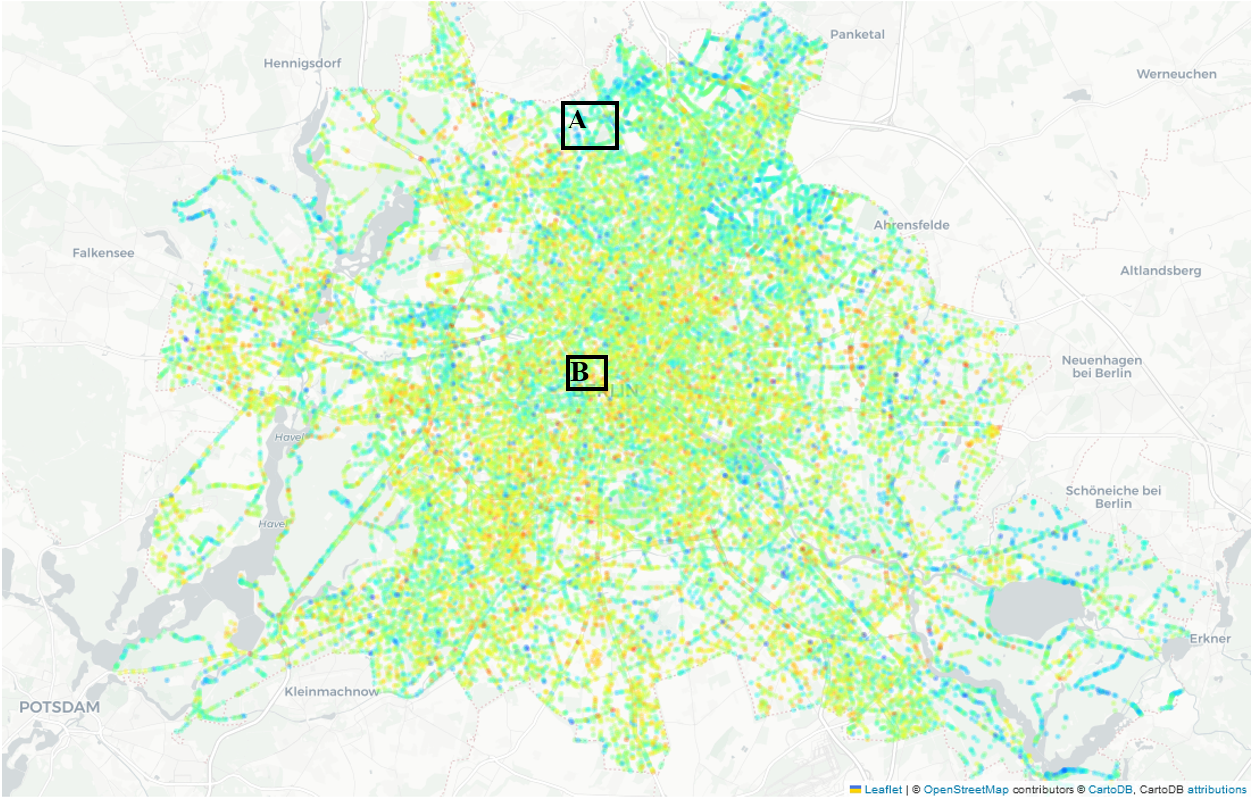}  \\
\end{minipage}
\begin{minipage}[b]{.09\textwidth}
    \centering
    \includegraphics[trim=550 100 380 100,clip, width=.98\textwidth]{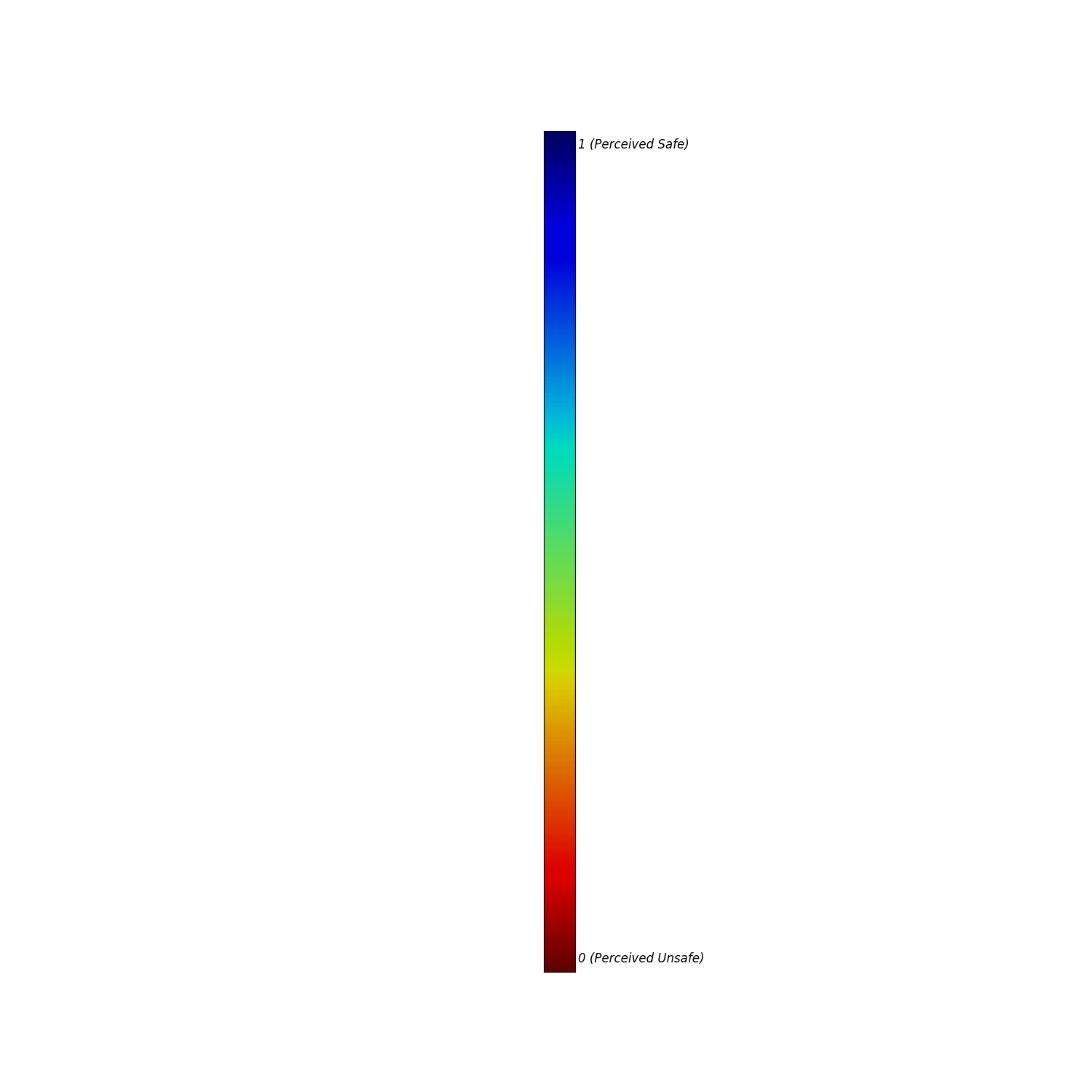}  \\
\end{minipage}
\hspace*{\fill}
\\\centering\small (a)

\vspace{6pt}
\begin{minipage}[b]{.49\textwidth}
    \centering
    \includegraphics[width=.98\textwidth]{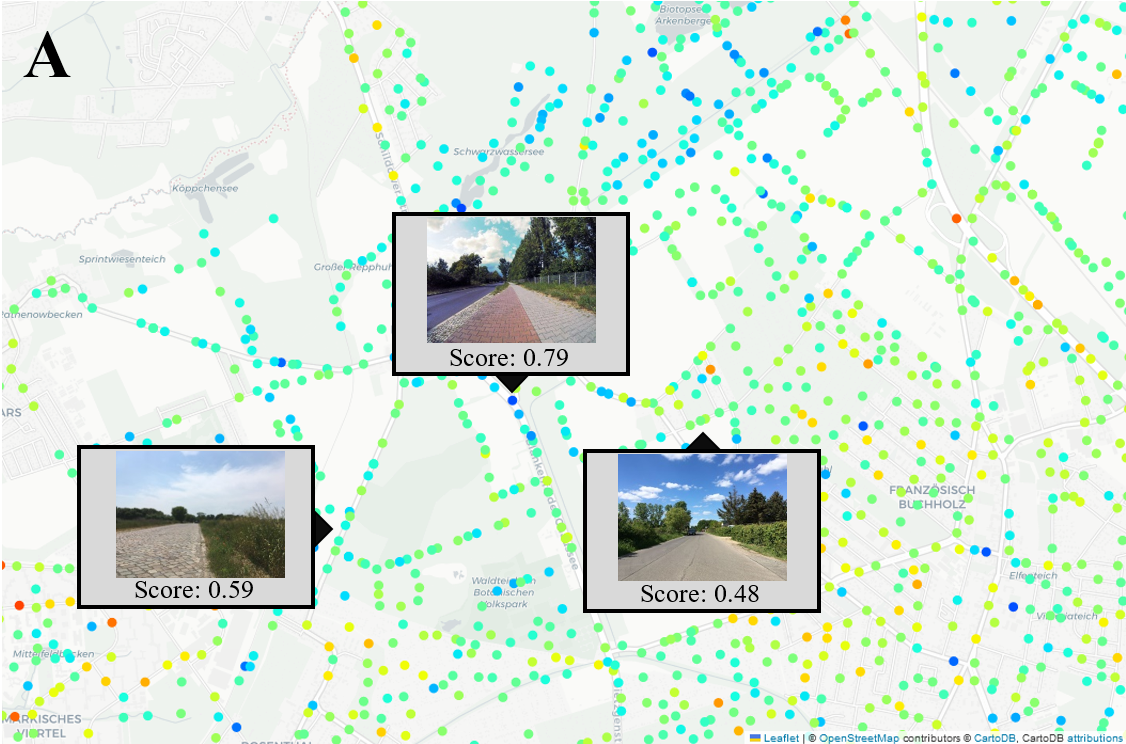}  \\
    \small (b)
\end{minipage}
\begin{minipage}[b]{.49\textwidth}
    \centering
    \includegraphics[width=.98\textwidth]{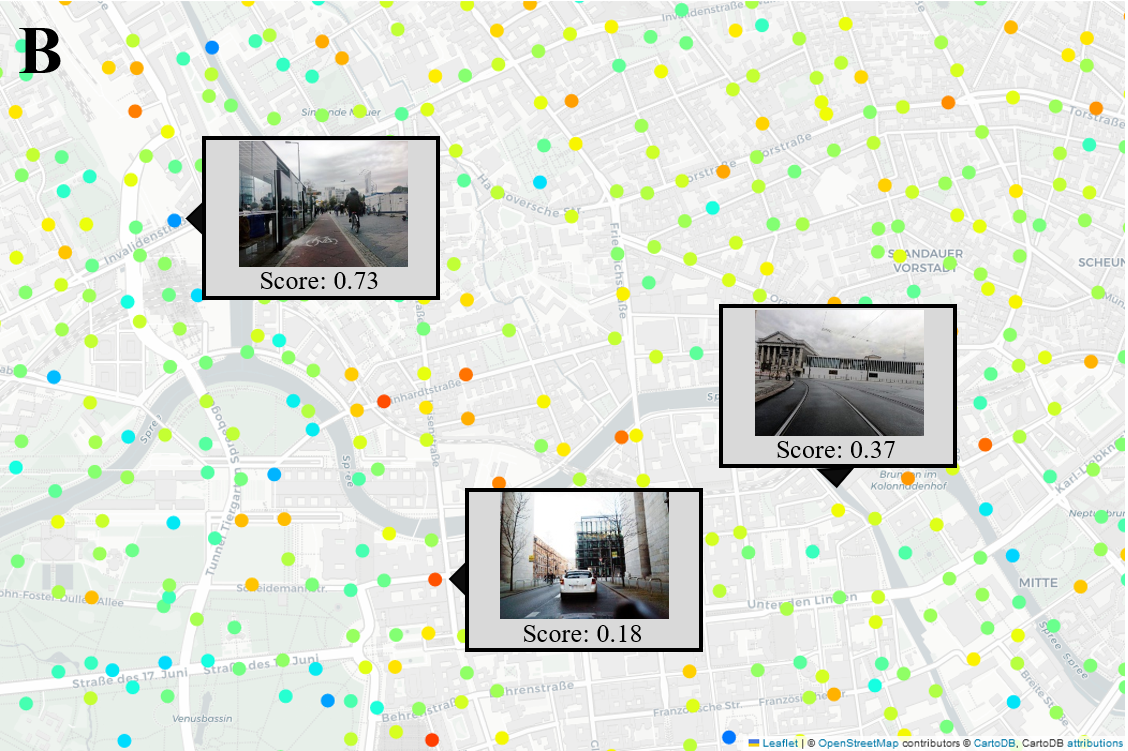}  \\
    \small (c)
\end{minipage}
  \centering
  \vspace{-3pt}
  \caption{Map of perceived cycling safety throughout Berlin (a). Scaled perceived safety scores, from safer (blue) to unsafer (red), are shown for sampled images in Berlin. Two areas are selected and shown in greater detail in (b) and (c).}
  \label{fig:app_map}
  \vspace{-6pt}
\end{figure*}

%\input{figs/application_details}

% ######################################################
% DATA
% ######################################################
To demonstrate how PCS-Net can be used to analyze the perceived cycling safety of an entire city, we computed perceived safety scores for the entire city of Berlin. To cover all of Berlin, we geographically sampled points using a 100m by 100m grid and projected them to the nearest cyclable path or road. We then retrieved street view images from Mapillary for each point. We extracted 36,700 unlabelled images, which we then ran through the most accurate PCS-Net model to extrapolate their perceived scores. This gave us a perceived cycling safety score for each image in Berlin, which, for readability, we scaled between [0, 1].

% ######################################################
% RESULTS
% ######################################################
Figure~\ref{fig:app_map} showcases the perceived safety scores for Berlin. As one would expect at the city-wide scale, scores are well distributed throughout the whole city. Taking a closer look (Figure~\ref{fig:app_map}.b), certain areas appear to be relatively perceived as safer (predominantly blue-ish, such as Berlin's north and southeast) and seem to be more continuous. Examining the corresponding images, we see that these correspond to less urbanized areas where high levels of vegetation are visible in the images. In the center of the city (Figure~\ref{fig:app_map}.c), locations are not perceived as safe as before, with scores varying widely within a relatively small area, even on locations on the same road/street which exhibit changes in the levels of perception. 

Additionally, we perform a human evaluations to ensure that PCS-Net estimated scores on previously unseen Berlin images accurately and it portraits the right level of perceived safety. To this end, a set of images with scores ranging from 0 (minimum score) to 1 (maximum score) were retrieved and shown to users to assess. Overall, environments with a lower score were consistently evaluated poorly in terms of safety, average-scored environments were neither considered very safe nor very unsafe, and environments whose scores were predicted higher were considered safer by our human evaluators.

On the whole, we showcase that there are several key hotspots for environments that are perceived as safe and unsafe, while at the same time, some high-level continuous hierarchy of perceived safety score seems to exist. From this, aggregate and disaggregate metrics can help planners and decision-makers highlight and prioritize urban environment changes to make cyclists' trips more enjoyable and provide them with more comfortable ones to cycle in.

\section{Discussion}
\label{sec:discussion}
% ######################################################
% DISCUSSION
% ######################################################
This work has focused on deriving cycling environments' perceived safety scores from pairwise comparisons. Pairwise image comparison surveys offer a systematic and quantitative approach to investigate visual preferences and individuals' perceptions, which can be used to quantify and categorize environments. Analyzing visual preferences from real-world images seeks to uncover the underlying patterns and principles that guide individuals' subjective perceptions. 

Unlike most traditional paired models that ignore or avoid tied comparisons between two items, here we have included and underlined the importance of including ties. This follows the idea of the seminal work of \cite{rao1967ties}, where ties matter and should be included to model comparisons when there is not enough difference between two items for a user's sense of perception to note a difference. In addition, allowing participants to choose a tie potentially increases their engagement, resulting in more truthful choices and more accurate models. Moreover, with reference to the Weber-Fechner laws of psychophysics, which explore stimulus magnitudes and the ability to distinguish between two stimuli, it would be interesting to model which environmental factors influence users' perceptions of cycling safety in such a way that they can distinguish between environments that appear safer for cycling. This knowledge could provide researchers with even more approaches to understanding the impact of different built environments on different cyclists.

All in all, much of the work developed warrants that ties should not be discarded when available. Results-wise, including them results in comparable performance to non-ties-only models, meaning that they can be included without any loss. Even with the limited number of average comparisons per image, we are able to derive good prediction power, meaning that we can learn directly from the presented images and survey responses to capture cycling safety perceptions. 

Another aspect that could be explored is the use of transformer-style networks, as these have recently gained popularity in computer vision tasks (namely object detection, semantic segmentation, and other learning tasks) for their higher prediction accuracies. Yet, these are characterized by requiring more data and processing power than traditional convolutional-style networks. It would, however, be interesting to explore the power of transformers and other network configurations as more data on the perception of cycling safety, together with street-view images, becomes available.

In practice, the approach we developed is extremely valuable not only in research but also in practice for urban planners and decision-makers to adequately address cyclists' needs in terms of sense of safety and comfort. New cities can employ a similar approach, deploying their own survey using specific city images and training a new PCS-Net specific to the city's needs and perceptions. In essence, this would help identify city areas (i.e., specific hotspots, streets, intersections, or whole neighbourhoods) where cyclists feel unsafer. In turn, this would allow city urban planners and transport authorities to more efficiently understand how the perception of cycling safety affects cyclists at that particular location and design changes to improve cyclists' sense of safety.

% ######################################################
% Limitations
\subsection{Limitations}

The analysis and model present in this work have explored how the perception of cycling safety and scoring cycling environments can be made using a siamese-convolutional neural network and a pairwise image comparison survey. Yet, it is important to also acknowledge its limitations, especially given the social and health aspects it indirectly covers.

First, it is important to consider that, despite our sample of images capturing various weather conditions, vehicle densities, lighting conditions, and road conditions, we could not account for all possible combinations and scenarios. For instance, while we included images with varying degrees of snow, vehicles (cars, bicycles, vans, trucks, buses) and pedestrians, images featuring sunny and cloudy depictions, as well as images during daytime, dawn, and dusk, other specific types of infrastructure that may limit visibility or alter cyclists' behaviour and more rare typologies of environments were not frequently present. Expanding and including a wider variety of environments in the survey may help address these gaps and enhance PCS-Net's validation across more contexts and environments. Directly accounting for specific external factors (e.g., weather conditions, different vehicles) is vital for urban planners and decision-makers to effectively create environments where cyclists feel safer. However, while such factors as weather were not directly controlled, from our results, we believe PCS-Net has indirectly been able to account for some of such factors shown directly on images. In practice, however, urban planners cannot change weather conditions; rather, they should consider their importance when designing and deploying new solutions. In turn, understanding how the built environment holistically may affect subjective safety may lead to more effective solutions to improve the sense of safety.

Second, we have demonstrated how PCS-Net can be used to extrapolate city-wide perceived scores. While this application is highly valuable for transport and road safety authorities, it is not without limitations. It is important to acknowledge that there is no definitive ground truth to compare the estimated scores or pairwise comparisons against, i.e. a sample of subjective evaluations collected on-site, where images were taken, to fully validate the use of images as surrogates of the real environment. We expect differences to exist since the experience of cycling in the real environment involves dynamic stimuli not present in an image evaluation experiment. Still, as documented in previous literature, these have been used as proxies (Section \ref{sec:related_work}) and future fieldwork will be required to close the gap in our specific context. Nevertheless, in our model validation, PCS-Net exhibited similar out-of-sample prediction accuracy to in-sample testing. While these highlight the potential of PCS-Net in estimating individuals' perception of safety, caution is warranted when applying PCS-Net to other contexts, such as different neighbourhoods, cities or countries where data was not available during training. Further analysis with larger and more comprehensive datasets is necessary to determine if the developed model can capture the general sense of safety perception across distinct and diverse settings. It is also important to assess whether PCS-Net can be used for knowledge transfer between cities with different cycling cultures and urban design philosophies.

Third, another important limitation regards our model's perception of safety scores when compared to \textit{in situ} perceptions cyclists may have. In this work, we have assumed the perception of safety to be captured directly from static images, which were correctly validated in themselves. However, we have not conducted any validation or comparison against safety perception data collected directly from cyclists on the road. Again, while PCS-Net presents itself as a novel and fast alternative to understanding the perception of safety at a large scale, care must be taken when directly assuming perceived scores on the road as PCS-Net does not account for urban dynamics or cyclists' own actions, behaviour and beliefs, which can also affect how cyclists perceived safety around them. Similarly, while this work tackles subjective safety, care must also be taken when relating it to objective safety, i.e., the number or severity of cycling accidents. The relation between subjective and objective safety is still not fully understood in contexts where both follow similar trends and others where perceived safety seems to be the opposite of actual safety \cite{winters2012safe, von2022crashvolume}, and thus caution must be exercised when trying to connect the two. Further work should analyse both how PCS-Net can be validated with \textit{in situ} interviews and how its scores relate to actual safety.

Fourth, to account for ties, we propose the use of margin $\gamma$ to model imperceptible differences between two images. We used a fixed $\gamma$, which, in more general terms, may differ across individuals. Yet, estimating independent $\gamma$'s for each user would allow for a better understanding of individual-level perceptions. Allowing for $\gamma$ to vary, either through bootstrapping or other approaches, may be used to ably model heterogeneity at an individual level or for different cycling profiles. Additionally, accounting for different cycling profiles and understanding specific differences between profiles can help decision-makers better tackle a group's needs more effectively, similar to what Guan et al. \cite{guan2021urban} do for incorporating regional interactions and attribute correlations.

\section{Conclusions}
\label{sec:conclusions}
% ######################################################
% CONCLUSIONS
% ######################################################
In this work, we have explored how cycling environment pictures can be ranked according to individuals' perceptions of safety. We base our work on pairwise comparisons, presenting participants with pairs of images and asking them to indicate their preferred choice of which environment appears safer for cycling. We then developed a siamese-style neural network that cannot only rank images based on choices (left or right image chosen), but also incorporate ties, often overlooked and ignored in the literature. Our proposed methodology achieved good results, requiring fewer observations than current paired models, as knowledge can be directly driven from image features and individuals' decisions. We extensively tested our approach on real-world data and real-world enriched data using synthetic images. Finally, we tested a city-wide application of our approach throughout Berlin.

% ######################################################
% FUTURE WORK
% ######################################################
In the future, we plan to use the approach detailed here to analyze other cities and understand if the perception of safety can be generalized across cities with different urban characteristics or cycling cultures. Moreover, not only is understanding what environments appear safer for cycling important, but knowing which of those environments' characteristics increase or decrease the sense of safety is vital. As such, we aim to identify urban elements and quantify their impact on the perception of safety in finer detail. 
One possible practical application where such information could be used is in the understanding of cyclists' route choices to verify how cyclists base their routes on safety perceptions.
This may help urban planners design environments where cyclists feel safer. 
Another possible path forward is the use of cycling videos instead of static images. This approach could help researchers understand how temporal changes and more urban dynamics impact the perception of cycling safety.
Finally, understanding the effect of individuals' cycling profiles and how they perceive cycling safety differently can also be valuable to better address individuals' needs.

\section*{ACKNOWLEDGMENTS}
This work is part of the research activity partially funded by Fundação para a Ciência e Tecnologia (FCT) via grant [PD/BD/142948/2018], supported by CERIS FCT funding [UIDB/04625/2020], LARSyS FCT funding (DOI: 10.54499/LA/P/0083/2020, 10.54499/UIDP/50009/2020, and 10.54499/UIDB/50009/2020), 
STREETS4ALL project [PTDC/ECI-TRA/3120/2021 (https://doi.org/10.54499/PTDC/ECI-TRA/3120/2021], 
PT Smart Retail project [PRR - 02/C05-i01.01/2022.PC645440011-00000062] through IAPMEI - Agência para a Competitividade e Inovação, and the Department of Technology, Management, and Economics at the Technical University of Denmark (DTU).

% % % % % % % % % % % % % % % % % % % % % % % % % % % % % % 
% Appendices
% % % % % % % % % % % % % % % % % % % % % % % % % % % % % % 

% % % % % % % % % % % % % % % % % % % % % % % % % % % % % % 
% References
% % % % % % % % % % % % % % % % % % % % % % % % % % % % % % 
\bibliographystyle{IEEEtran}
\bibliography{IEEEabrv, references}

% % % % % % % % % % % % % % % % % % % % % % % % % % % % % % 
% BIO
% % % % % % % % % % % % % % % % % % % % % % % % % % % % % % 
\vskip -3\baselineskip plus -1fil
\begin{IEEEbiographynophoto}{Miguel Costa}
is a postdoc researcher at the Department of Technology, Management, and Economics of the Technical University of Denmark (DTU). He holds a Ph.D. (2024) in Transportation Systems and a MSc in Electrical and Computer Engineering from Instituto Superior Técnico (IST), University of Lisbon. His research focuses on understanding and predicting how cyclists' objective and subjective safety is influenced by the context where cyclists cycle and the surrounding built environment and how to use reinforcement learning for finding optimal climate adaptation policies by integrating dynamic links between flooding events, transport, health, and wellbeing.
\end{IEEEbiographynophoto}
\vskip -3\baselineskip plus -1fil
\begin{IEEEbiographynophoto}{Manuel Marques}
is a Researcher in the Electrical and Computer Engineering Department at IST, and also at the Instituto de Sistemas e Robótica, LARSyS. He holds a Ph.D. in Electrical and Computer Engineering from IST (2011) and his main area of research is computer vision with a special interest in 3D reconstruction, video processing object recognition, and pose estimation. Recently, Dr. Marques focused on applications to assistive robotics and transportation, in particular, cycling mobility.
\end{IEEEbiographynophoto}
\vskip -3\baselineskip plus -1fil
\begin{IEEEbiographynophoto}{Carlos Lima Azevedo}
is an Associate Professor at the Department of Technology, Management, and Economics at DTU and a Research Affiliate of the ITSLab, Massachusetts Institute of Technology (MIT). He has a Ph.D. (2014), MSc (2008), and 5y-BSc in Structural Engineering (2004) all from U. of Lisbon. He is an expert in the development of new models and simulation techniques for smart mobility design and assessment, which includes developing and applying large-scale agent-based urban simulation for the design and evaluation of shared automated vehicles on-demand and personalized real-time incentive systems; individual behaviour experiments and preferences modelling towards new mobility solutions, and innovative road safety assessment models.
\end{IEEEbiographynophoto}
\vskip -3\baselineskip plus -1fil
\begin{IEEEbiographynophoto}{Felix Wilhelm Siebert}
is an Assistant Professor at the Human Behavior Section at the Department of Technology, Management, and Economics at DTU. He holds a Ph.D. in Transport Psychology. His main research interest is the safety-related behavior of vulnerable road users. In current work, he applies computer vision approaches to video data for the automated detection of safety-related behavior of road users.
\end{IEEEbiographynophoto}
\vskip -3\baselineskip plus -1fil
\begin{IEEEbiographynophoto}{Filipe Moura}
is an Associate Professor with Habilitation in Transportation of the Department of Civil Engineering and Architecture at IST. He is head of the U-Shift lab of the Transport Systems Research Group at CERIS. His main research expertise is on Urban Mobility, focusing on analyzing travel behavior, transport modeling, and sustainable mobility. His endeavors emphasize active modes and their interactions within the urban environment and explore autonomy within urban mobility (from both behavioral and technological perspectives). Filipe is distinguished as a Fulbright Scholar.
\end{IEEEbiographynophoto}

\end{document}